\documentclass[runningheads]{llncs}
\usepackage[T1]{fontenc}
\usepackage{graphicx}
\usepackage{subfigure}
\usepackage{booktabs}
\usepackage[misc]{ifsym}
\newcommand{\corr}{(\Letter)}
\usepackage{todonotes}
\usepackage{amsmath}
\usepackage{multirow}
\usepackage{hyperref}
\usepackage{mdframed}
\usepackage{orcidlink}
\begin{document}

\title{Variance-Aware Noisy Training: Hardening DNNs against Unstable Analog Computations}

\titlerunning{Variance-Aware Noisy Training}

\author{Xiao Wang\orcidlink{0009-0001-5064-7940}\corr \and Hendrik Borras\orcidlink{0000-0002-2411-2416} \and
	Bernhard Klein\orcidlink{0000-0003-0497-5748} \and
	Holger Fröning\orcidlink{0000-0001-9562-0680}}
\authorrunning{X. Wang et al.}

\institute{Hardware and Artificial Intelligence Lab, Institute of Computer Engineering, Heidelberg University, Germany\\
\email{\{xiao.wang,hendrik.borras,bernhard.klein,holger.froening\}@ziti.uni-heidelberg.de}}
\maketitle              %

\begin{abstract}
The disparity between the computational demands of deep learning and the capabilities of compute hardware is expanding drastically.
Although deep learning achieves remarkable performance in countless tasks, its escalating requirements for computational power and energy consumption surpass the sustainable limits of even specialized neural processing units, including the Apple Neural Engine and NVIDIA TensorCores.
This challenge is intensified by the slowdown in CMOS scaling.

Analog computing presents a promising alternative, offering substantial improvements in energy efficiency by directly manipulating physical quantities such as current, voltage, charge, or photons.
However, it is inherently vulnerable to manufacturing variations, nonlinearities, and noise, leading to degraded prediction accuracy.
One of the most effective techniques for enhancing robustness, Noisy Training, introduces noise during the training phase to reinforce the model against disturbances encountered during inference.
Although highly effective, its performance degrades in real-world environments where noise characteristics fluctuate due to external factors such as temperature variations and temporal drift.

This study underscores the necessity of Noisy Training while revealing its fundamental limitations in the presence of dynamic noise.
To address these challenges, we propose Variance-Aware Noisy Training, a novel approach that mitigates performance degradation by incorporating noise schedules which emulate the evolving noise conditions encountered during inference.
Our method substantially improves model robustness, without training overhead.
Through experiments on image classification tasks in dynamic noise environments, we demonstrate a significant increase in robustness, from 79.3\% with conventional Noisy Training to 97.6\% with Variance-Aware Noisy Training on CIFAR-10 and from 32.4\% to 99.7\% on Tiny ImageNet.

\keywords{noisy training  \and noisy computations \and analog computing \and robustness \and neural networks.}
\end{abstract}

\newpage
\section{Introduction}
\label{sec:intro}

Deep neural networks (DNNs) have driven remarkable advancements in a wide array of machine learning applications, from computer vision and natural language, speech and signal processing.
These breakthroughs are largely enabled by digital compute platforms, such as graphics processing units (GPUs) or specialized accelerators, which offer high throughput and flexibility. 
However, as DNNs grow in scale and are increasingly deployed in energy-constrained environments, the quest for more efficient hardware solutions becomes paramount. 
In addition, Complementary Metal-Oxide-Semiconductor (CMOS) technology scaling is stuttering, thus alternative approaches to maintain performance scaling have to be found. 
Analog computing architectures replace discrete quantities with continuous ones, leveraging the inherent properties of physical systems to perform computations efficiently.  
These architectures enhance computational efficiency, reduce data movement overhead, and enable highly parallel multiply-accumulate (MAC) operations, while their most significant advantage lies in superior energy efficiency~\cite{chi2016prime,shafiee2016isaac}. 
Analog accelerators leverage the intrinsic physical properties of existing and emerging device technologies—such as analog CMOS-based computing \cite{BSS-2}, photonic computing~\cite{Feldmann2021}, resistive random-access memory (ReRAM), phase-change memory (PCM), and other non-volatile devices—to perform approximate MAC operations based on physical quantities such as charge, even directly in the memory array \cite{yang2013memristive}. 
These architectures can significantly cut down power consumption and latency, surpassing many of their digital counterparts \cite{Burr02012017,Ambrogio2018}. 
However, these advantages come at the cost of increased susceptibility to analog non-idealities and noise. 
Factors such as device variations, thermal fluctuations, mismatch, drift, and aging can degrade both the performance and reliability of analog DNN implementations \cite{shafiee2016isaac,gokmen2016acceleration,Sebastian2020}.

From a machine learning perspective, various works have reported that adding small amounts of noise to the training data can improve generalization \cite{bishop1995trainingNoise,deep-learning,Holmstrom1992addnoise}, thus acting as a form of data augmentation. Noise injection is often also referred to as “distortion” or “jitter”, in particular in early works. 
Besides injecting such (usually Gaussian) noise to input variables, there are similar methods on adding noise to other parts of a neural architecture, including weights \cite{deep-learning}, gradients \cite{Neelakantan2015} and activations \cite{gulcehre2016noisyAF}. 
However, this applies only to small noise levels and is limited to training, while inference remains noise-free.
Thus analog hardware noise has the potential to distort intermediate activations and weights, undermining the model's inference accuracy if left unmitigated. 
Consequently, there is substantial interest in techniques that preserve DNN accuracy under noise.  
One of the most widely studied methods is Noisy Training, where noise is intentionally injected during the training process to emulate the hardware imperfections encountered during inference \cite{noisy_machine,he2019noiseinjection}. 
Exposure to noise from the outset enables the model to adapt its parameters, enhancing robustness to real-world noise variations.  
From a broader perspective, there are adjacent works in multiple directions: on adversarial effects to improve the robustness of DNNs \cite{Tsipras2018}, on Noisy Training to introduce sparsity in the activation space \cite{bricken2023sparsenoise}, as well as on using noise in physical computations as a source of stochasticity \cite{brckerhoffplckelmann2024probabilistic}.
While Noisy Training has been shown to be successful in various use cases, it is important to note that its efficacy strongly depends on the fidelity with which one can replicate the true hardware noise characteristics during training. 
When there is a mismatch---e.g., in distribution (statistical shape), amplitude (magnitude), or temporal correlation (noise in real hardware sometimes changes over time)---between training noise and inference noise, DNNs often fail to generalize the learned robustness and may suffer a decrease in prediction quality.
It is important to emphasize that in analog devices, maintaining a constant noise level is highly uncommon. 
Even under controlled laboratory conditions, stabilizing noise over time presents significant challenges. 
Environmental factors such as temperature fluctuations, electromagnetic induction, and various timing effects inherently influence noise characteristics. 
Given these dynamic influences, it is reasonable to assume that noise in analog hardware is not static but evolves over the device's operational lifetime. 
This variability underscores the necessity of considering time-dependent noise models when designing robust deep learning systems for analog accelerators.
This observation raises an important research question: How precisely must one capture the analog hardware's noise characteristics during training to ensure robust inference? 
Addressing this question is non-trivial, given that the noise profiles in analog circuits can evolve over time due to changing environmental conditions, device aging, or even variations in operating modes.

A second challenge arises when perfectly matching the real hardware noise in training is either impractical or impossible. 
While techniques like hardware-in-the-loop training can provide more accurate noise profiles \cite{neftci2019surrogateSNN}, they may be expensive or time-consuming to implement. 
In practical implementations, only approximations or partial knowledge of noise statistics may be available.  
This raises a key question: How can training algorithms ensure robustness when training noise only partially matches deployment conditions?  

In response to these challenges, research has increasingly focused on rigorous noise modeling and robust training schemes that can handle real-world analog non-idealities \cite{walkingnoise,bernhard-incremental,Kuhn2023,neftci2019surrogateSNN,noisy_machine,he2019noiseinjection}. 
Understanding how noise affects different DNN layers and accumulates through network depth is crucial for mitigating its impact.  
Sophisticated training techniques—ranging from gradient-based noise modeling to Bayesian approaches—have been proposed to enhance model reliability under noisy conditions \cite{noisy_machine}. 
Ultimately, the goal is to facilitate a new generation of DNN accelerators that can achieve cutting-edge performance while maintaining a significantly lower energy footprint.
In this paper, we systematically explore the interplay between Noisy Training and real-world dynamic noise environments. Specifically:

\begin{itemize}
\item \textbf{Quantifying Noise Mismatch:} We study how varying degrees of mismatch between the noise injected during training and the real noise encountered in inference affect the final model accuracy.%

\item \textbf{Evaluation of Robustness Techniques:} We evaluate the effectiveness of robustness techniques, including Noisy Training, Quantization, and Perturbation on weights, in dynamic noise environments characteristic of analog hardware. Our results indicate that Noisy Training significantly outperforms the alternative methods, establishing it as the most reliable baseline for robustness in such settings.

\item \textbf{Strategies for Imperfect Noise Knowledge:} 
We propose a novel training procedure designed to mitigate the adverse effects of partial or inaccurate noise assumptions: \textit{Variance-Aware Noisy Training (VANT)}%

\item \textbf{Guidelines for Robust Analog DNN Training:} Drawing on our theoretical and experimental findings, we offer practical guidelines on how to tailor \textit{VANT} to diverse analog hardware setups.%
\end{itemize}

By addressing these facets, we contribute to the broader effort of understanding and optimizing DNNs for analog hardware deployment. 
Our results demonstrate that carefully designed Noisy Training enables robust energy-efficient inference, even under non-ideal and time-varying hardware noise.  
Ultimately, we aim to offer both theoretical and practical contributions that inform the design of next-generation hardening methods for DNNs on analog accelerators.

\section{Related Work}
\label{sec:related}

Neural network robustness is a critical research area, addressing threats such as adversarial attacks, compression errors, and computational noise.  
Noise injection plays a central role both as an evaluation metric and a training technique to enhance resilience.  
This section overviews research on robustness, noise injection, and Noisy Training in analog computing.  

\paragraph{Quantization and Robustness}

Quantization, a prevalent technique in hardware-efficient deep learning, reduces numerical precision and thereby affects model robustness.
Prior research has extensively examined its impact on adversarial resilience.
For example, studies have demonstrated that adversarial robustness exhibits a non-monotonic relationship with bit-width, indicating that increased precision does not always enhance robustness \cite{quant_bits_adv}.
Similarly, findings suggest that quantization can improve resilience against adversarial attacks while incurring minimal accuracy loss \cite{quant_adv_relative_robustness}.
To mitigate error amplification that exacerbate adversarial perturbations, methods such as controlling the Lipschitz constant during quantization have been proposed \cite{defensive_quant}.
Further analyses have investigated quantization effects across various neural architectures, revealing that highly complex models can recover from severe weight quantization through retraining, whereas smaller models experience greater performance degradation \cite{Resiliency_of_QNN}.

\paragraph{Perturbation and Robustness}\label{para:perturbation-robustness}
By perturbing the weights of a DNN, SGD can find regions within the parameter space, which are more robust in general.
Or inversely: When injecting noise into a DNN, its predictions are more likely to be correct if the DNN is trained towards a robust loss region.
Motivated by the relationship between the loss landscape sharpness and generalization, SAM~\cite{sam} seeks parameter values whose entire neighborhoods maintain consistently low training loss. Moreover, other research has incorporated pertutbation on both weights and inputs, improving the robustness against adversarial attacks~\cite{awp}.
\paragraph{Noise Injection for Robustness}\label{para: noise-injection-adversarial}

Noise injection has long been recognized as an effective strategy to improve generalization in machine learning models \cite{murray1994,grandvalet1997noiseinjection}.
Early works explored its utility in mitigating overparameterization, comparing it with techniques such as weight decay and early stopping \cite{jiang2009effectNI}.
With the rise of adversarial attacks, noise injection evolved into a robust defense mechanism alongside adversarial training \cite{goodfellow2015adversarialtraining}.
Various approaches have been proposed, including globally injected additive Gaussian noise \cite{He2019parametricNoise} and ensembles leveraging layer-wise noise injection \cite{Liu2018selfensemble}.
However, they often assume static noise distributions, overlooking dynamic variations in real-world scenarios.  

\paragraph{Noisy Training in Analog Computing}

Unlike adversarial perturbations that affect input sensitivity, noisy analog hardware primarily introduces stochasticity into internal computations, particularly affecting neural network weights and dot product calculations.
Studies have modeled non-volatile memory noise as an additive zero-mean i.i.d. Gaussian noise term on model weights, demonstrating the benefits of injecting similar noise during training and extending robustness via knowledge distillation \cite{noisy_machine}.
Other research has incorporated memristor perturbation models to simulate drift in neural network weights, capturing long-term instability in analog devices and proposing architecture search and layer-specific dropout to increase robustness against drifts \cite{noisy_bayes}.

Dependent on the considered hardware implementation noise might be dominant in different parts of the accelerator and thus occur at different positions in the computations.
Techniques have been developed to address noise from both weight readout~\cite{noisy_machine,noisy_bayes} and subsequent computations, such as injecting noise at the output activation level \cite{walkingnoise}.
Thereby, latter accounts for accumulated noise and extends further by introducing layer-specific noise to evaluate robustness and learning dynamics~\cite{walkingnoise}.  

Additionally, Noisy Training approaches have been extended to exploit noise as an inherent feature of analog computing systems, enhancing adversarial robustness and supporting stochastic inference \cite{cappelli2022,wu2022harnessing,brckerhoffplckelmann2024probabilistic}.

Despite their effectiveness, most Noisy Training strategies assume static noise characteristics.  
This limits real-world deployment, where noise fluctuates due to temperature changes, voltage instability, and device aging.  
Variance-Aware Noisy Training addresses this by integrating dynamic noise schedules that reflect realistic inference-time variations.  

\section{Neural Networks and Noisy Environments}
We begin with an overview of the datasets and model architectures used in our work, followed by a comprehensive analysis of existing methods and their characteristics.  
Additionally, we describe the methodology for simulating a noisy analog environment and outline our approach to evaluating robustness.  

\paragraph{Datasets, Models and Experimental Setup}
In order to establish the effectiveness of our proposed method, experiments are performed for different networks and datasets.
For the initial and comprehensive evaluation, we perform image classification on CIFAR-10~\cite{cifar-10}, CINIC-10~\cite{CINIC10} and Tiny ImageNet~\cite{Le2015TinyIV}.
For CIFAR-10, we evaluate two model architectures: LeNet-5~\cite{lenet5} and ResNet-18~\cite{resnet}.
While for Tiny ImageNet and CINIC-10, LeNet-5 is undersized and thus we focus on ResNet-18 and ResNet-50. 

LeNet-5 and ResNet-18 use initial learning rates of 0.001 and 0.01 on CIFAR-10, respectively, while both ResNet-18 and ResNet-50 use a learning rate of 0.001 on Tiny ImageNet and CINIC-10.  
All models are trained with Adam and cosine learning rate decay, using a batch size of 128 for 400 epochs. \footnote{The code is available at: \url{https://github.com/HAWAIILAB/VANT}}

\subsection{Global Noise Injection and Noisy Training}\label{sec:background-global-noise}
\begin{figure}[hbtp]
	\centering
	\includegraphics[width=0.95\textwidth]{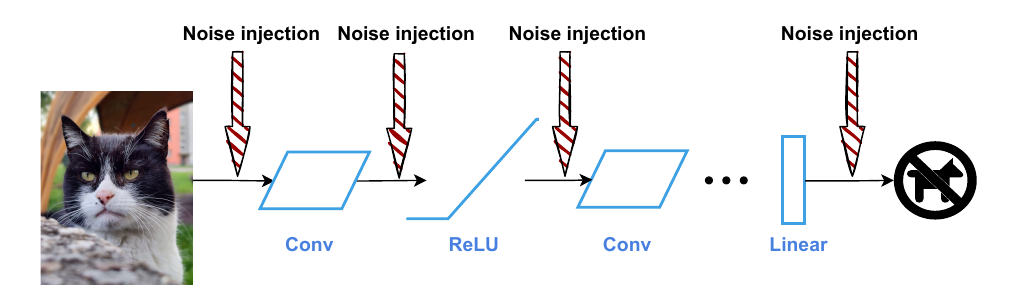}
	\caption{Global noise injection in a DNN. Noise is applied to activations between layers.}  
	\label{fig:global-noise-injection}
\end{figure}
To simulate the noisy environment present in analog hardware, we inject noise at a global level during model computation.  
In this work, we follow the \textit{Walking Noise}~\cite{walkingnoise} methodology, which focuses on injecting noise at the activations. 
We consider additive Gaussian noise, due to its widespread occurrence in natural processes and its demonstrated effectiveness in previous works on noise injection~\cite{noisy_machine}. To inject noise without bias, we sample with zero mean, i.e. $\mathcal{N} \left( 0, \sigma\right)$, with $\sigma$ being the standard deviation of the noise.  
A schematic of the noise injection is shown in Figure~\ref{fig:global-noise-injection}.  
By varying the noise level \(\sigma\) during inference, we assess the model's robustness under noisy environments of different intensities.  

\begin{figure}[htp]
	\centering
	\includegraphics[width=0.775\linewidth]{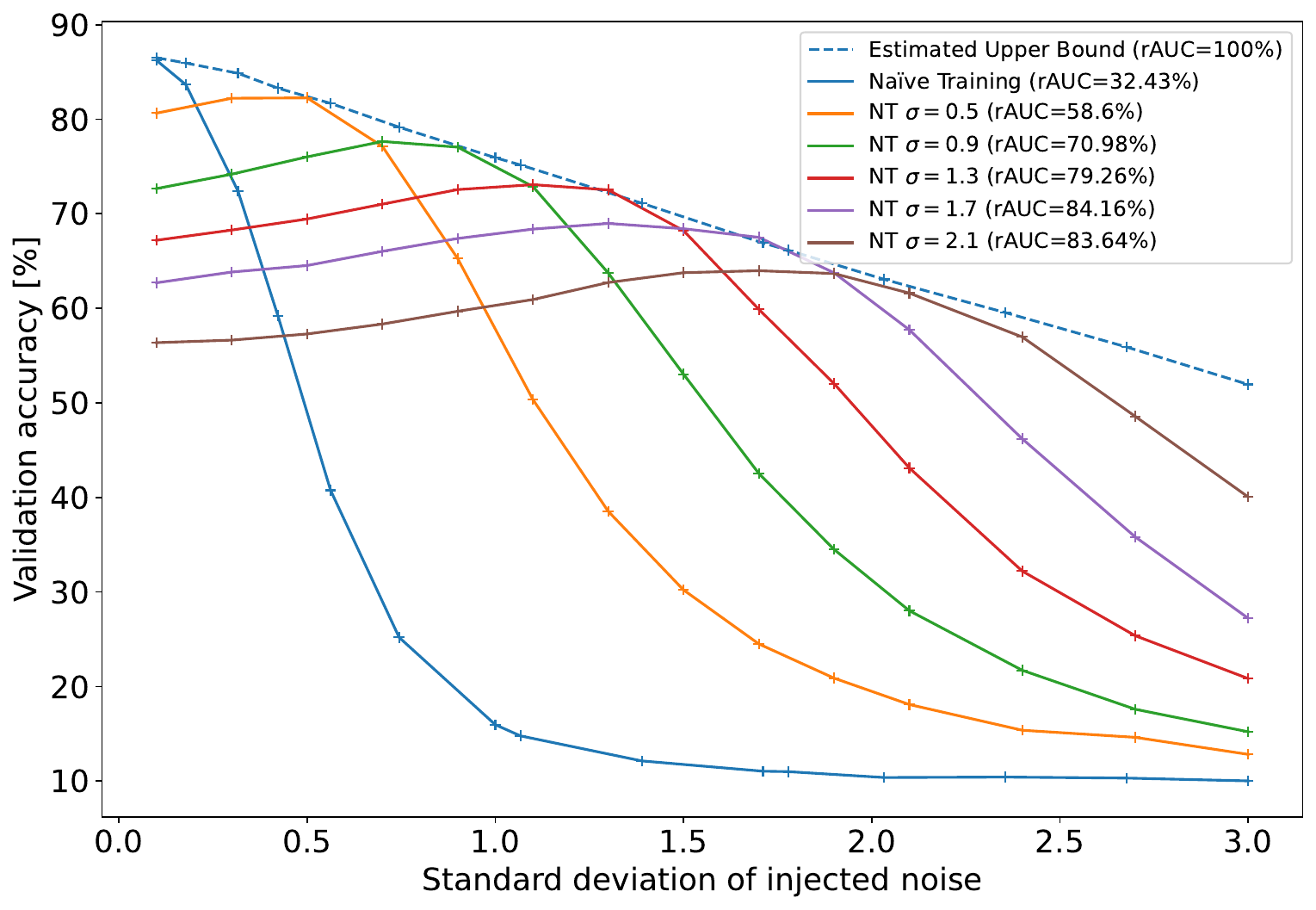}
	\caption{Accuracy degradation under noise. NT: Models trained with Noisy Training (NT) at noise of $\sigma$. Dashed line indicates the upper bound, when using the same noise for training and inference. ResNet-18 on CIFAR-10.}\label{fig:resnet-basic}
\end{figure}

Noise injection during training has been shown to significantly improve network accuracy under noisy computations~\cite{bernhard-incremental,noisy_machine}.  
We also evaluate performance of standard Noisy Training by injecting noise in the forward pass during the training procedure. 
Figure~\ref{fig:resnet-basic} reveals the impact of noise injection to accuracy for models trained with and without noise injection. 
When a model is trained with the same noise level it encounters during inference, it typically achieves optimal accuracy. 
By connecting these optimal points, we obtain the dashed curve, which represents the best achievable performance using Noisy Training at each noise level. 
We assume the dashed curve thus to be the theoretical upper bound in terms of robustness and accuracy for any given noise level.

\subsection{Evaluation: Quantifying Robustness under Noise}\label{sec:background-evaluation}
The standard deviation of the noise $\sigma$ under test is selected from the range $[0.1, 3.0]$. This selection is based on previous findings~\cite{bernhard-incremental}, which report noise levels on analog hardware to fall within this interval. This ensures that our robustness assessment is both relevant and practical for deployment scenarios involving noisy analog computations.

To quantify the robustness under noisy computation, we utilize the Area Under the Curve (AUC) as primary performance metric. 
However, directly comparing AUC values can be misleading, as accuracy is influenced by factors such as model complexity and dataset difficulty. Moreover, absolute AUC values do not directly indicate how close a method's performance is to the upper bound.
To provide a fair comparison, we use the \textbf{relative AUC percentage (rAUC)}, defined as:
$$\text{rAUC}=\frac{\text{AUC of the method}}{\text{AUC of the upper bound curve}} [ \% ]$$
This metric directly indicates how close a method is to the best possible result. 

\section{Noisy Training: Strong but Not Flawless}
As shown, training with noise injection significantly enhances robustness compared to a baseline model trained without noise. However, it is important to understand how other robustness-enhancing methods contribute to overall performance.
In the following, we explore these methods as well as the limitations of Noisy Training.
\begin{figure}[htp]
	\centering
	\includegraphics[width=0.7\linewidth]{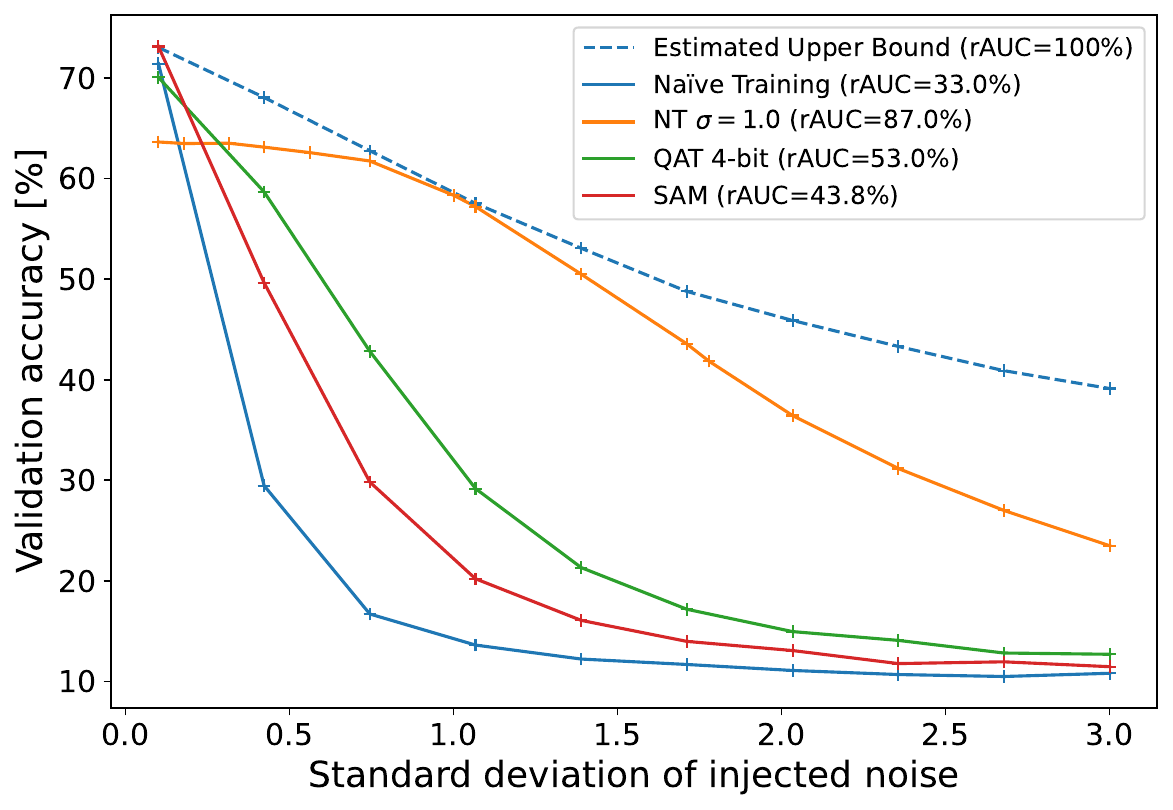}
	\caption{Comparison of typical hardening methods for LeNet-5 on CIFAR-10.}\label{fig:lenet-comp}
\end{figure}
\subsection{The Importance of Noisy Training}
We begin with the assumption that we do not have prior knowledge of noise characteristics in analog hardware and explore how to mitigate its impact. We consider quantization and the generalization method Sharpness-Aware Minimization (SAM)~\cite{sam} as potential countermeasures.

Intuitively, quantization introduces quantization error during computation, which may however increase the stability of DNNs when subjected to noise perturbations. 
To evaluate the impact of quantization techniques on the robustness of neural networks against computational noise, we test a quantized network in a noisy environment.
For our evaluations we employ quantization-aware training (QAT)\footnote{Our implementation is based on Brevitas (https://github.com/Xilinx/brevitas)}. Figure~\ref{fig:lenet-comp} shows that a model quantized to 4-bit outperforms the baseline model.
We further evaluate the impact of the perturbation method SAM, as described in section \ref{sec:related}.
As shown in Figure~\ref{fig:lenet-comp}, while SAM offers improvements it remains less effective than quantization. 

However, neither quantization nor SAM can match the performance of Noisy Training in enhancing robustness. This is largely due to the fundamental characteristics of noise in analog hardware: it is pervasive, affecting not only the input but also internal computations; its magnitude can be substantial; and, critically, it accumulates as signals propagate through the neural network.
These factors highlight that an effective countermeasures must account for the specific noise properties of the hardware. 

\subsection{Limitations of Noisy Training}
\label{sec:limitation-noisy-training}
While Noisy Training is essential for robustness, a key challenge remains: the noise characteristics of analog hardware can fluctuate %
over time due to environmental factors such as temperature variations. Additionally, different hardware units usually exhibit notable variations in noise levels. We define the noise level present in a specific hardware instance at the time of measurement as $ \sigma_\text{train}$.

This raises an important question: even if a model is trained under a specific noise level, how well does it generalize when the on-device noise deviates from the training conditions? To explore this, we train the model under a fixed noise level and then evaluate its performance across different noise strengths. The orange curve in Figure~\ref{fig:lenet-comp} illustrates the performance of LeNet-5 trained with $ \sigma_\text{train}=1.0$. As expected, the model achieves optimal performance when the noise level matches the training condition. However, as the noise deviates from $ \sigma_\text{train}=1.0$, accuracy declines, %
highlighting sensitivity to mismatched noise levels.

This observation leads to a crucial conclusion: Noisy Training is only effective when the noise characteristics are precisely known. If the noise level during training does not align with the actual noise encountered during deployment, the model's robustness can be significantly compromised. 
This leads to the central research question of this work: 
\begin{mdframed}[backgroundcolor=gray!10, linecolor=black, linewidth=1pt, roundcorner=5pt]
\centering
How can we train models that remain robust across an entire fleet of devices, each potentially exhibiting different noise strengths over time?
\end{mdframed}
\section{Beating the odds: Variance-Aware Noisy Training}

In order to address the previously presented shortcomings of Noisy Training we present a novel training technique, \textit{Variance-Aware Noisy Training (VANT)} which is more robust against unstable noise settings.

\subsection{Methodology: Variance-Aware Noisy Training}

The central assumption behind standard (stable) Noisy Training is that one can model the accelerator's noise perfectly, in particular, that it will remain constant over time and devices.
This way gradient descent adjusts a given DNN to the characteristics of a given accelerator.
Centrally missing however is any treatment of variation in the noise.
We thus extend Noisy Training as follows:
\begin{equation} \label{eq:var_of_var}
	\begin{aligned}
		x &\sim \mathcal{N} \left( 0, \sigma_\text{var} \right),\\
		\sigma_\text{var} &\sim \mathcal{N} \left( \alpha \cdot \sigma_\text{train}, \theta \right).
	\end{aligned}
\end{equation}
Here $\sigma_\text{train}$ is an extrinsic parameter, representing the known noise characteristic of a given hardware target.
\textit{VANT} additionally introduces two parameters: 
$\theta$ adjusts Noisy Training to the time variations of a given accelerator, while
$\alpha$ is a calibration parameter for $\sigma_\text{train}$.
Looking forward to Sections \ref{sec:exp:cifar10} and \ref{sec:exp:TinyImageNet}, we note that \textit{VANT} is rather insensitive to $\alpha$, while $\theta$ strongly depends on the chosen $\sigma_\text{train}$.

During training $\sigma_\text{var}$ is then sampled for each input image, while additively injected noise ($x$) is sampled for each activation.
All sampling and thus noise injection only applies during the forward pass of gradient descent training.

\subsection{Experiments on CIFAR-10}
\label{sec:exp:cifar10}

In order to evaluate how the parameters of \textit{VANT} behave, we run initial evaluations on CIFAR-10. 
In a later step we then evaluate how well \textit{VANT} transfers to a more complex dataset, when utilizing the same parameters obtained here.

\begin{figure*}[hbtp]
	\centering
	\subfigure[Constant $\alpha=0.3$]{\includegraphics[width=0.495\textwidth]{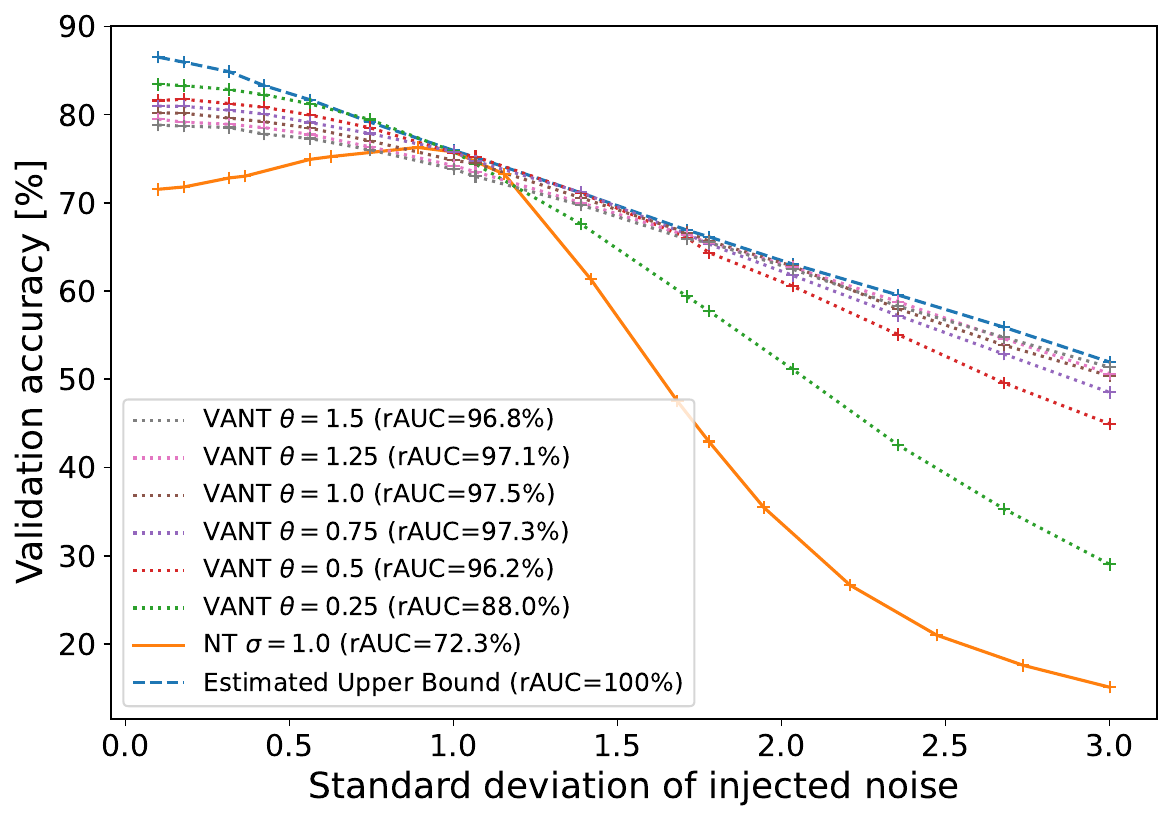}\label{fig:cifar-10-theta-alpha-individual_a}}
	\subfigure[Constant $\theta=0.75$]{\includegraphics[width=0.495\textwidth]{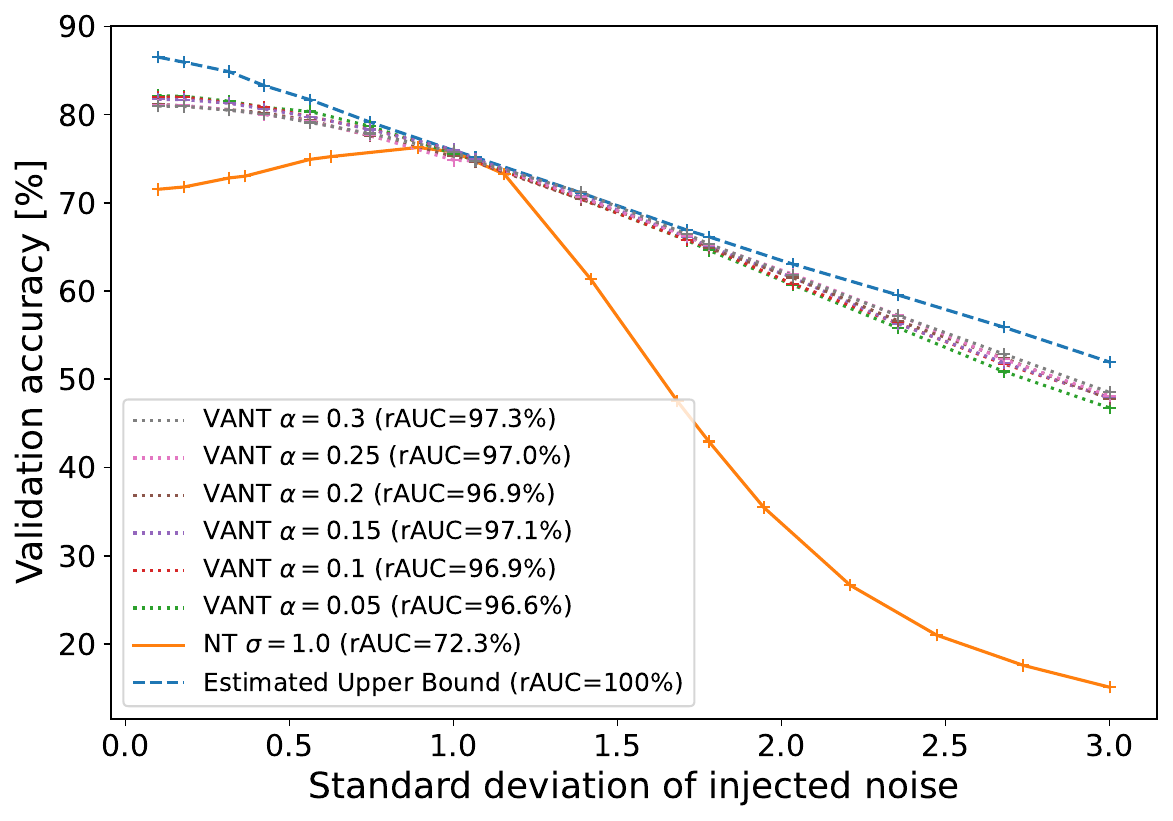}\label{fig:cifar-10-theta-alpha-individual_b}}
	\caption{Exploration of \textit{VANT} hyperparameters for ResNet-18 on CIFAR-10, when plotted over the injected noise. Here compared to Noisy Training (NT) at $\sigma_\text{train}=1.0$.}\label{fig:resnet18-cifar10}
	\label{fig:cifar-10-theta-alpha-individual}
\end{figure*}

Initially we evaluate $\alpha$ and $\theta$ individually to explore their general behavior (Figure \ref{fig:cifar-10-theta-alpha-individual}). 
For any setting of $\alpha$ and $\theta$, the robustness (rAUC) of \textit{VANT} is significantly better compared to standard Noisy Training.
And results are similar for LeNet-5.
While variations in $\alpha$ appear to have little impact on the overall robustness as seen in Figure \ref{fig:cifar-10-theta-alpha-individual_b}, $\theta$ plays a significant role in both the overall robustness and shape of the curve, see Figure \ref{fig:cifar-10-theta-alpha-individual_a}.
Furthermore, in Figure \ref{fig:cifar-10-theta-alpha-individual_a} it also becomes apparent that the method does not necessarily preserve the peak accuracy of standard Noisy Training at $\sigma_\text{train}$.
The variation of noise, $\theta$, effectively increases the maximum noise observed, which shifts the point of optimal accuracy for \textit{VANT}.
Thus, during the evaluation of $\alpha$ and $\theta$  this influence needs to be considered. 
This is accomplished by measuring how close the accuracy at $\sigma_\text{train}$ is to standard Noisy Training, since ideally \textit{VANT} should preserve all advantages of Noisy Training.
In order to quantify this behavior, a new metric is introduced: \textit{Preserved Accuracy}.
It measures by how much the accuracy of \textit{VANT} and standard Noisy Training differ at the noise injection point of $\sigma_\text{train}$.
Ideally a setting of $\alpha$ and $\theta$ can be found, which keeps this metric at zero or higher, perfectly preserving the accuracy of standard Noisy Training.

\begin{figure*}[hbtp]
	\centering
	\subfigure[Robustness (rAUC \%)]{\includegraphics[height=3.75cm]{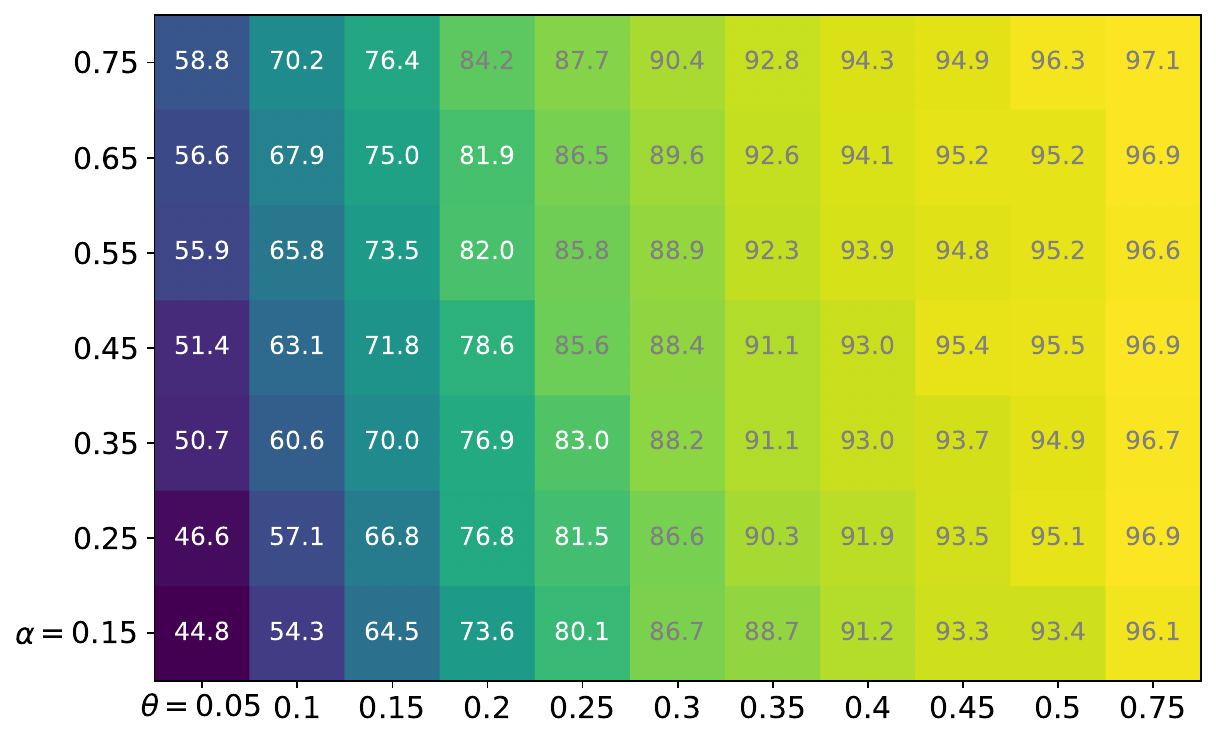}\label{fig:heatmap-resnet18-cifar_a}}
	\subfigure[Preserved Accuracy]{\includegraphics[height=3.75cm]{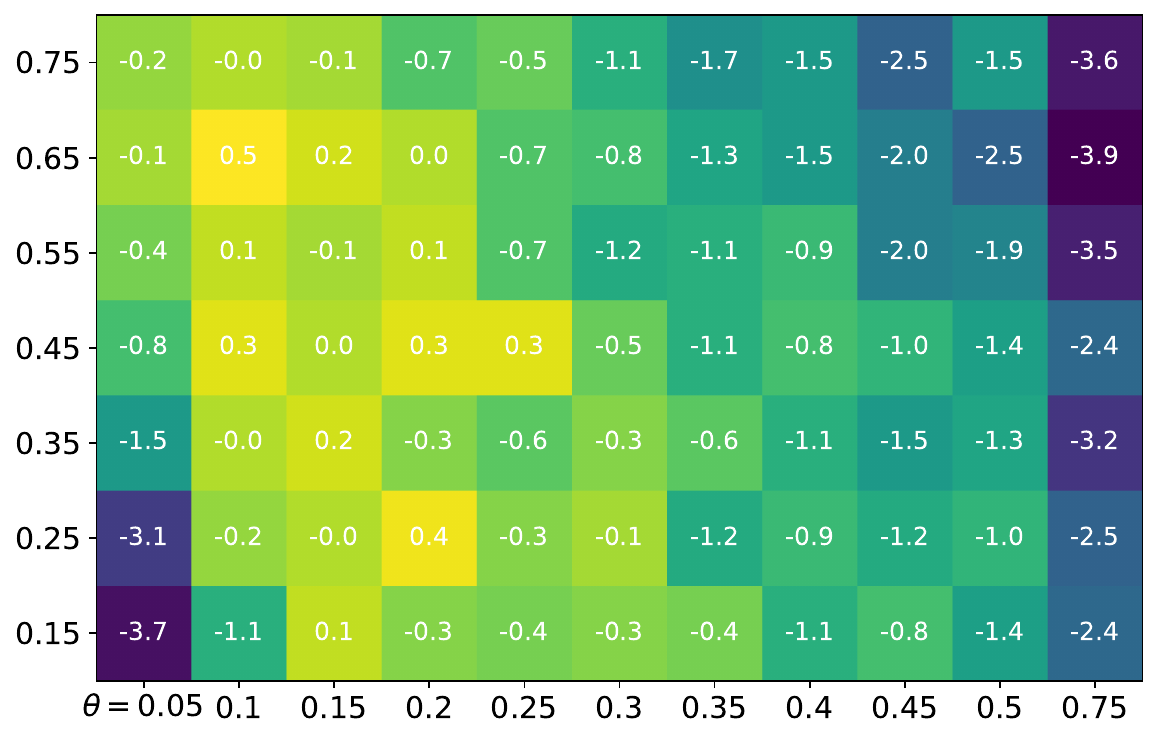}\label{fig:heatmap-resnet18-cifar_b}}
	\caption{Heatmap of quality metrics for \textit{VANT} with ResNet-18 on CIFAR-10, when varying both hyperparameters of \textit{VANT}, while keeping $\sigma_\text{train}=0.4$ constant.}
	\label{fig:heatmap-resnet18-cifar}
\end{figure*}

To further identify the dependency of the robustness (rAUC) and preserved accuracy on $\alpha$ and $\theta$, a grid scan for both parameters across a wide range is performed in Figure \ref{fig:heatmap-resnet18-cifar}.
For the robustness we primarily observe that $\theta$ plays a strictly monotonic role in improving robustness.
We further note that $\alpha$ similarly monotonically increases the robustness. %
While this effect appears to suggest that one should simply increase both $\alpha$ and $\theta$, this is not the case.
Instead the maximally achievable robustness is bounded by the preserved accuracy, as this value should stay at zero or larger in Figure \ref{fig:heatmap-resnet18-cifar_b}.
Notably a sweet spot becomes visible for %
finding an optimal set of parameters for \textit{VANT}.
While this sweet spot is well constrained in $\theta$, it is relatively broad for $\alpha$.

Selecting the best set of $\alpha$ and $\theta$ parameters is done as follows: 
\begin{enumerate}
	\item Select all sets of $\alpha$ and $\theta$, for which the preserved accuracy is above 0, ensuring parity to Noisy Training.
	\item Sub-select $\alpha$ and $\theta$ for which the robustness is maximized.
\end{enumerate}
In the case of Figure \ref{fig:heatmap-resnet18-cifar}, we select $\alpha=0.45$ and $\theta=0.25$ as the optimal parameters.

Since different analog accelerators require different initial noise levels, we now explore how \textit{VANT} behaves for different $\sigma_\text{train}$. %
As \textit{VANT} is largely invariant to $\alpha$ we fix it to $0.45$,  as a middle ground for the sweet spot from Figure \ref{fig:heatmap-resnet18-cifar_b}.
\begin{figure*}[bt]
	\centering
	\subfigure[Robustness (rAUC \%)]{\includegraphics[width=0.485\textwidth]{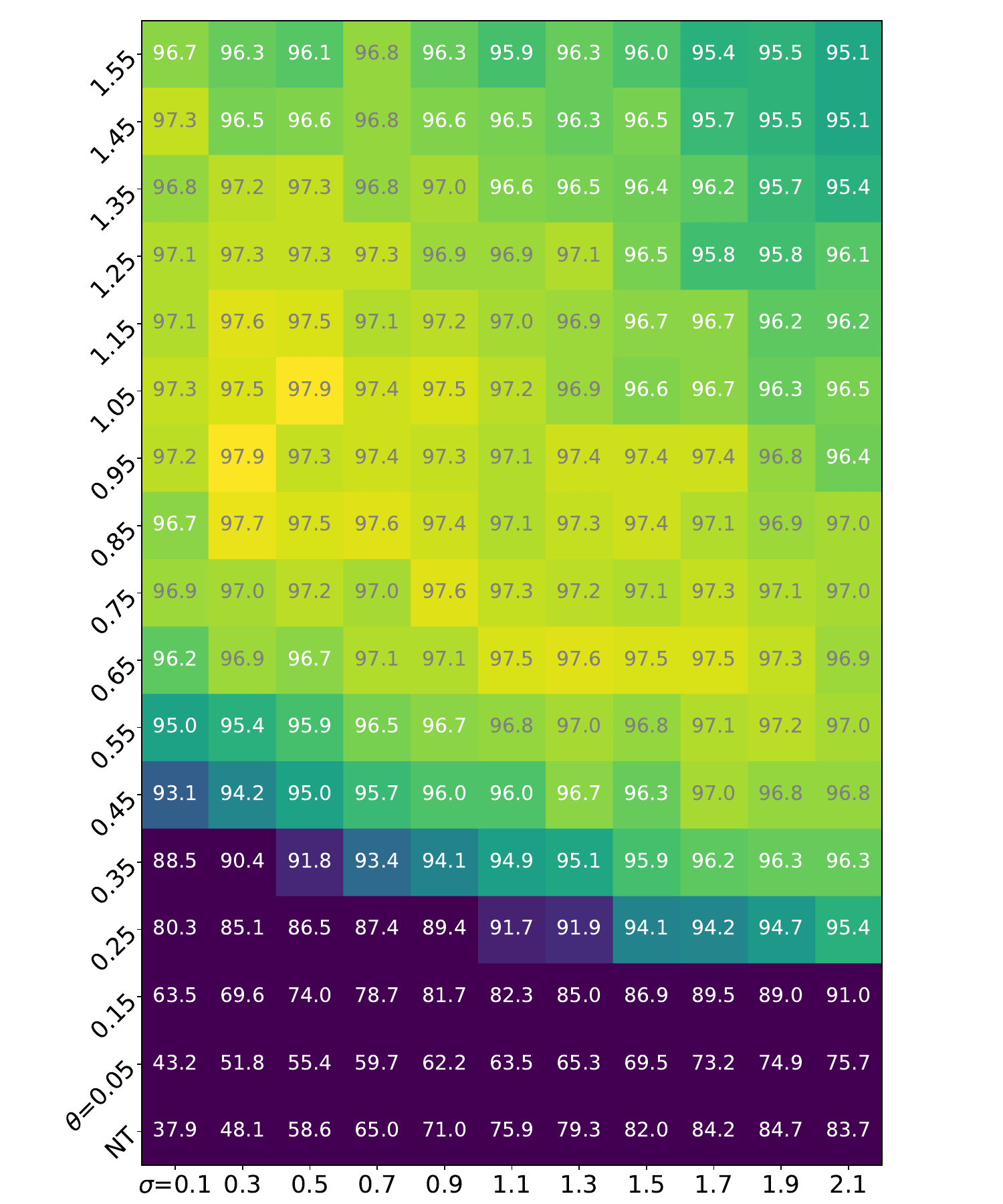}}
	\subfigure[Preserved Accuracy]{\includegraphics[width=0.485\textwidth]{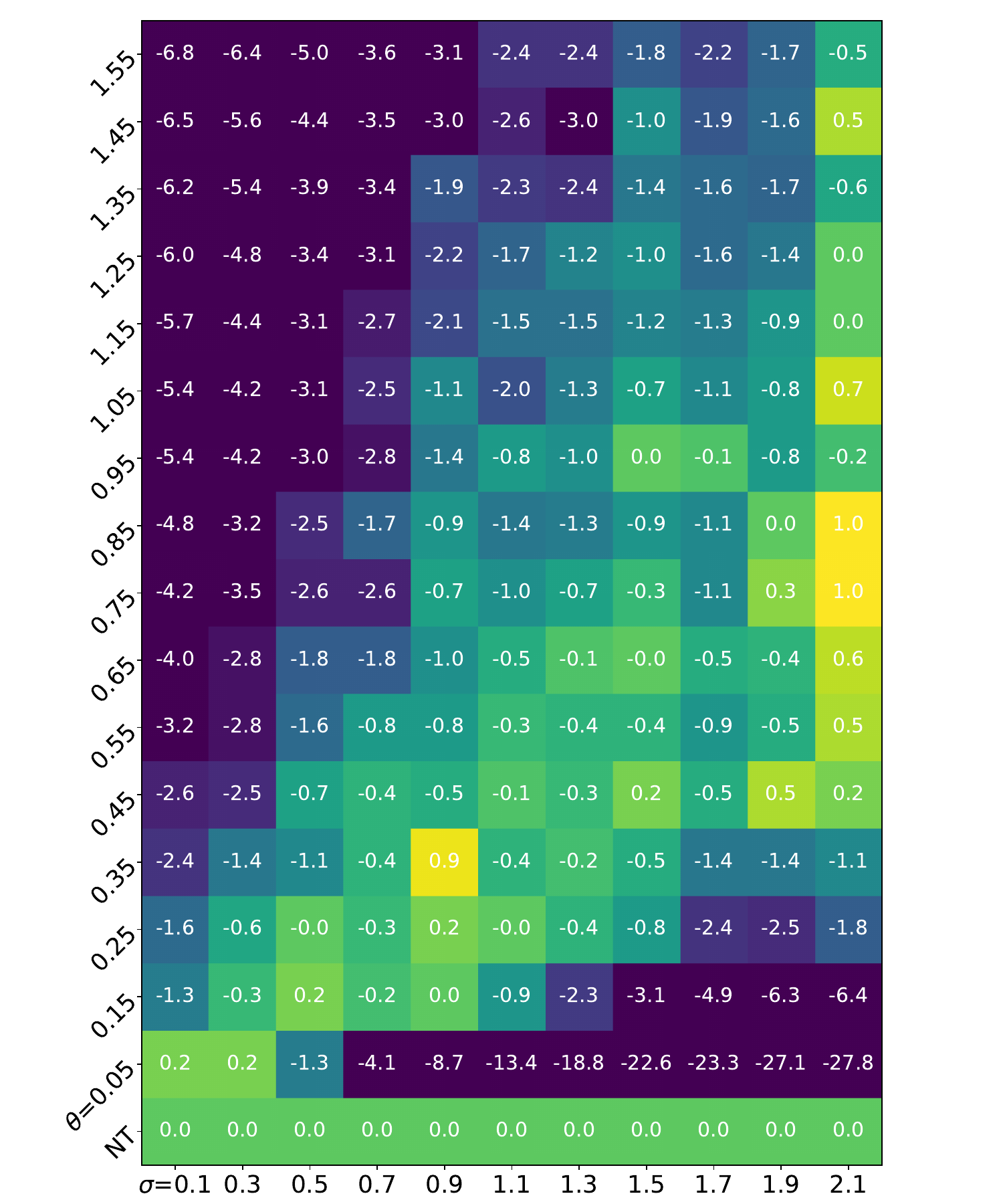}}
	\caption{Heatmap of quality metrics for \textit{VANT} with ResNet-18 on CIFAR-10, when varying $\theta$ and the reference hardware noise $\sigma_\text{train}$. While keeping $\alpha=0.45$ constant, as it is largely invariant. NT: Noisy Training as baseline at the bottom.}
	\label{fig:heatmap-resnet18-fix-alpha}
\end{figure*}

Similarly to Figure \ref{fig:heatmap-resnet18-cifar}, Figure \ref{fig:heatmap-resnet18-fix-alpha} shows both robustness and preserved accuracy.
However, in this case the x-axis denotes the change in $\sigma_\text{train}$, e.g. different hardware accelerators, and the y-axis explores the behavior of $\theta$.
Again, there is a trade-off to be made between the preserved accuracy and the robustness.

Notably, $\theta$ shows a broad optimum for the robustness.
However, when following the procedure for selecting the best $\theta$ as stated in the steps above, then $\theta$ is tightly confined by an approximately linear relationship between $\sigma_\text{train}$ and $\theta$, which is approximately: $\theta = 0.4 \cdot \sigma_\text{train}$
\begin{table}[]
	\centering
	\caption{Quality metrics for \textit{VANT} with ResNet-18 on CIFAR-10 for the optimal $\theta$ and $\alpha$, when varying the reference hardware noise $\sigma_\text{train}$.} \label{tab:optimal-theta}
	\setlength{\tabcolsep}{0.35em} %
	{\renewcommand{\arraystretch}{1.2}%
	\begin{tabular}{c|c |c|c|c|c|c|c|c|c|c|c}
	
		$\sigma_\text{train}$ & 0.1 &0.3 & 0.5&0.7 & 0.9 & 1.1 & 1.3 & 1.5  & 1.7  & 1.9 & 2.1  \\ \hline \hline
		$\theta$ & 0.05 & 0.05 & 0.25 & 0.35 & 0.35 & 0.45 & 0.65 & 0.65 & 0.95 & 0.75 & 0.85 \\ \hline
		Preserved Accuracy &  0.2 & 0.2& 0 & -0.4& 0.9 & -0.1 & -0.1 & 0 & -0.1& 0.3 & 1.0 \\ \hline
		rAUC [\%] NT & 37.9\ &48.1 &58.6 &65.0 & 71.0& 75.9& 79.3 &82.0 &84.2 &84.7 &83.7 \\ \hline
		rAUC [\%] VANT & 43.2\ &51.8 &86.5& 93.4& 94.1& 96.0& \textbf{97.6} &97.5 &97.5 &97.1 &97.0 \\
	\end{tabular}
	}
\end{table}

However, for all following experiments, we choose $\theta$ as described with the steps above, as these are more accurate given the ground truth data available. 
These can be found in Table \ref{tab:optimal-theta}.
All selected settings for \textit{VANT} simultaneously preserves the accuracy of standard Noisy Training and strictly improve the robustness, as visible by comparing them to the bottom row in Figure \ref{fig:heatmap-resnet18-fix-alpha}.

\subsection{Generalizing to Complex Data: CINIC-10 \& Tiny ImageNet}
\label{sec:exp:TinyImageNet}
In the following we explicitly test for two properties of \textit{VANT}:
How well it transfers to more complex datasets and if the parameters of \textit{VANT} additionally show stability across varying architectures.
As such all settings for $\alpha$ and $\theta$ for the following experiments are chosen from the optimal results of the previous section, shown in Table \ref{tab:optimal-theta}.
\begin{figure*}[hbtp]
	\centering
	\subfigure[Robustness (rAUC)]{\includegraphics[width=0.485\textwidth]{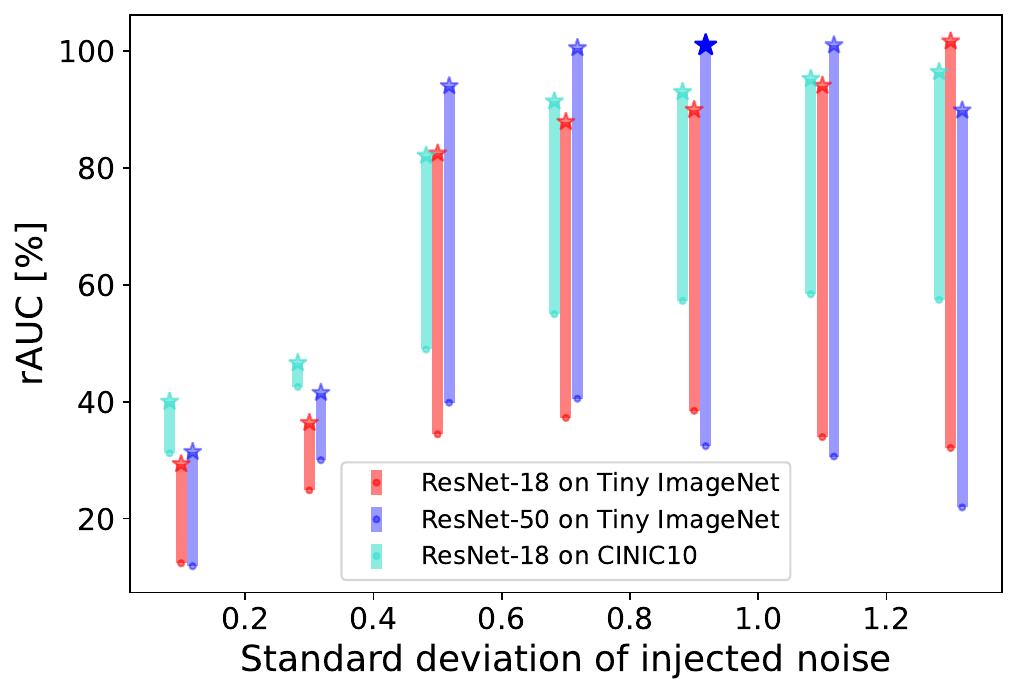}\label{fig:cinic-tinyImageNet_a}}
	\subfigure[Preserved Accuracy]{\includegraphics[width=0.485\textwidth]{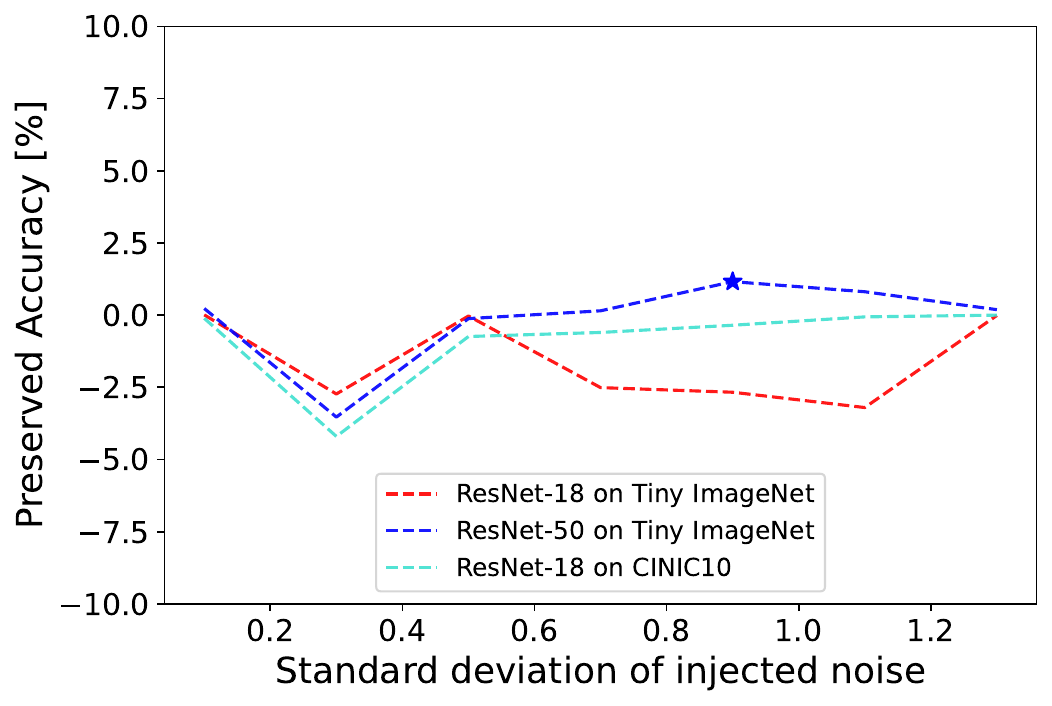}\label{fig:cinic-tinyImageNet_b}}
	\caption{Quality metrics for \textit{VANT} with ResNet-18/50 on CINIC-10 and Tiny ImageNet at varying $\sigma_\text{train}$. For the robustness (rAUC): The dot at the bottom of the bars represents standard Noisy Training, while the star at the top of the bar represents \textit{VANT}.}
	\label{fig:cinic-tinyImageNet}
\end{figure*}

As a naturally more complex architecture we investigate ResNet-50. 
On the dataset side we increase complexity in two steps: CINIC-10 and Tiny ImageNet.%

Results for both quality metrics under consideration are shown in Figure \ref{fig:cinic-tinyImageNet}.
Looking first at the preserved accuracy, one can observe that the accuracy is generally preserved across datasets and models.
A notable exception is $\sigma=0.3$, where the preserved accuracy drops across all experiments and increases in robustness are also low.
We postulate that for this specific setting $\theta$ was ill-chosen.
Further investigating the robustness in Figure \ref{fig:cinic-tinyImageNet_a}, we observe that \textit{VANT} improves the robustness for all models, datasets and strengths of injected noise.
However the effect is not consistent across the whole range of injected noise. 
\textit{VANT} provides less significant robustness improvements for noise strengths of $\sigma_\text{train} < 0.5$.
The underlying reason is that in order to achieve high accuracy under low noise, regions of higher noise are less prominently sampled. 
This results in subpar performance for much of the investigated range by the rAUC metric.
Notably this behavior is largely inherited from Noisy Training.

For $\sigma_\text{train}\ge 0.5$, however, \textit{VANT} is considerably more robust.
Interestingly the largest improvements can be found on Tiny ImageNet with ResNet-50, the most complex dataset and model investigated, where the rAUC increases dramatically from 32.4\% to 99.7\% at $\sigma_\text{train}=0.9$.
Nonetheless, siginificant improvements are also visible for ResNet-18 and CINIC-10.

Concluding we find, that \textit{VANT} shows good generalization across both datasets and models. 
While the performance of a set of chosen hyper-parameters remains consistent at the same time.

\section{Summary}
\label{sec:summary}

The increasing computational demands of modern deep learning models pose significant challenges for conventional digital CMOS technology, which is approaching fundamental scaling limits. 
To address this bottleneck, alternative computing paradigms have gained attention, with analog computing emerging as a promising candidate. 
However, analog accelerators introduce new challenges, particularly due to inherent noise that can degrade model performance.

In this work, we first examine the challenges associated with training DNNs under these imperfect conditions. 
We observe that while techniques such as quantization and SAM contribute to improved model robustness, they fall short of the robustness provided by Noisy Training. 
However, Noisy Training itself has critical limitations: although it can achieve high peak accuracy, it exhibits poor generalization when subjected to variations in noise levels, as typically encountered in analog hardware. 
Specifically, standard Noisy Training tends to overfit to a particular noise configuration, leading to suboptimal performance when the noise characteristics shift due to factors such as temperature fluctuations and hardware aging—common occurrences in analog accelerators.

To address this robustness gap, we propose \textit{Variance-Aware Noisy Training (VANT)}, an extension of standard Noisy Training that explicitly accounts for temporal variations in noise. 
\textit{VANT} incorporates an additional term which models the expected evolution of the noise environment over time. 
It thus enhances the generalization capabilities of DNNs under real-world deployment conditions, where noise characteristics are dynamic rather than fixed.

Empirical evaluations demonstrate the effectiveness of \textit{VANT} in improving robustness across different noise regimes and dataset complexities. 
For instance, under typical analog noise conditions, \textit{VANT} increases robustness from 79.3\% to 97.6\% on CIFAR-10. 
On the more challenging Tiny ImageNet dataset, VANT similarly yields significant gains, improving performance from 32.4\% to 99.7\%.

In summary, our findings highlight a crucial principle for deploying DNNs on noisy analog hardware:
it is not sufficient to account solely for the immediate noise environment during training; rather, it is essential to model the temporal evolution of noise over time.
By adopting a more comprehensive approach that considers the dynamic nature of hardware noise, \textit{VANT} represents a significant step toward enabling robust deep learning models on fleets of analog accelerators.

\begin{credits}
	\subsubsection{\discintname}
	The authors have no competing interests to declare that are
	relevant to the content of this article.
\end{credits}

\bibliographystyle{splncs04}
\bibliography{references}

\end{document}